\documentclass[conference]{IEEEtran}
\usepackage{times}

\usepackage[numbers]{natbib}
\usepackage{multicol}
\usepackage{graphics,graphicx,caption,float,subcaption,booktabs,xcolor,multirow,array,color,ifthen,tabu,colortbl,dblfloatfix,url,xparse,mathtools,patchcmd,algorithm,algorithmic,amssymb,xspace,nicefrac,microtype,amsmath,amsfonts,bm,ragged2e,tikz,stackengine,etoolbox,xpatch,enumerate,xstring,setspace,tabularx,makecell,changepage,cuted,titlesec,wrapfig,tcolorbox, sidecap,comment}
\usepackage[pagebackref=true,breaklinks=true,colorlinks=true,bookmarks=false,citecolor=blue]{hyperref}
\usepackage{adjustbox}
\usepackage[english]{babel}
\usepackage[normalem]{ulem}

\usepackage[symbol]{footmisc}

\begin{document}

\title{Robotic Telekinesis: Learning a Robotic Hand Imitator by Watching Humans on YouTube}
\author{Aravind Sivakumar\footnotemark*$\qquad$Kenneth Shaw\footnotemark*$\qquad$Deepak Pathak\\Carnegie Mellon University}

\newcommand{\baseline}[0]{DexPilot-Monocular$^*$}
\newcommand{\shortbaseline}[0]{DexPilot-Mono$^*$}

\maketitle
 
\begin{abstract}
We build a system that enables any human to control a robot hand and arm, simply by demonstrating motions with their own hand. The robot observes the human operator via a \textit{single RGB camera} and imitates their actions \textit{in real-time}. Human hands and robot hands differ in shape, size, and joint structure, and performing this translation from a single uncalibrated camera is a highly underconstrained problem. Moreover, the retargeted trajectories must effectively execute tasks on a physical robot, which requires them to be temporally smooth and free of self-collisions. Our key insight is that while paired human-robot correspondence data is expensive to collect, the internet contains a massive corpus of rich and diverse human hand videos. We leverage this data to train a system that understands human hands and retargets a human video stream into a robot hand-arm trajectory that is smooth, swift, safe, and semantically similar to the guiding demonstration. We demonstrate that it enables previously untrained people to teleoperate a robot on various dexterous manipulation tasks. Our low-cost, glove-free, marker-free remote teleoperation system makes robot teaching more accessible and we hope that it can aid robots  {in learning} to act autonomously in the real world.  {Video demos} can be found at: \href{https://robotic-telekinesis.github.io}{https://robotic-telekinesis.github.io}
\end{abstract}
\footnotetext[1]{Equal Contributions}

\IEEEpeerreviewmaketitle

\section{Introduction}
{Mimicking human behavior with robots} has been a central component of robotics research for decades.  This paradigm, known as teleoperation, has successfully been used to enable robots to perform tasks that were unsafe or impossible for humans to perform, such as handling nuclear materials \cite{vertut1985robot} or deactivating explosives \cite{davies2001technology}. Teleoperation has also been used to enable the robotic automation of tasks that are easy for humans to demonstrate but difficult to program.  In industrial robotics, for example, teleoperation can be used to demonstrate a single trajectory (e.g. picking a box from a conveyor belt) that the robot \textit{overfits} to {and repeats} verbatim for months or years thereafter.  Teleoperation can alternatively be used as a means to collect a large dataset of demonstrations, which can then be used to learn a policy that \textit{generalizes} to new tasks in unseen environments \cite{rajeswaran2017learning, radosavovic2020state}.

 In this paper, we specifically study the problem of teleoperation for dexterous robotic manipulation. While there are many promising current techniques (e.g. Kinesthetic Control \cite{billard2006discriminative}, Virtual Reality devices \cite{oculus, vive}, haptic gloves \cite{HaptX} and MoCap \cite{zhao2012combining}), each {of them} suffers {from} some shortcoming that has limited its applicability.  These setups typically involve expensive hardware and specialized engineering, expert operators, or an apparatus that impedes the natural fluid motion of the demonstrator’s hand.
\begin{figure}[t]
 \centering
 \includegraphics[width=\linewidth]{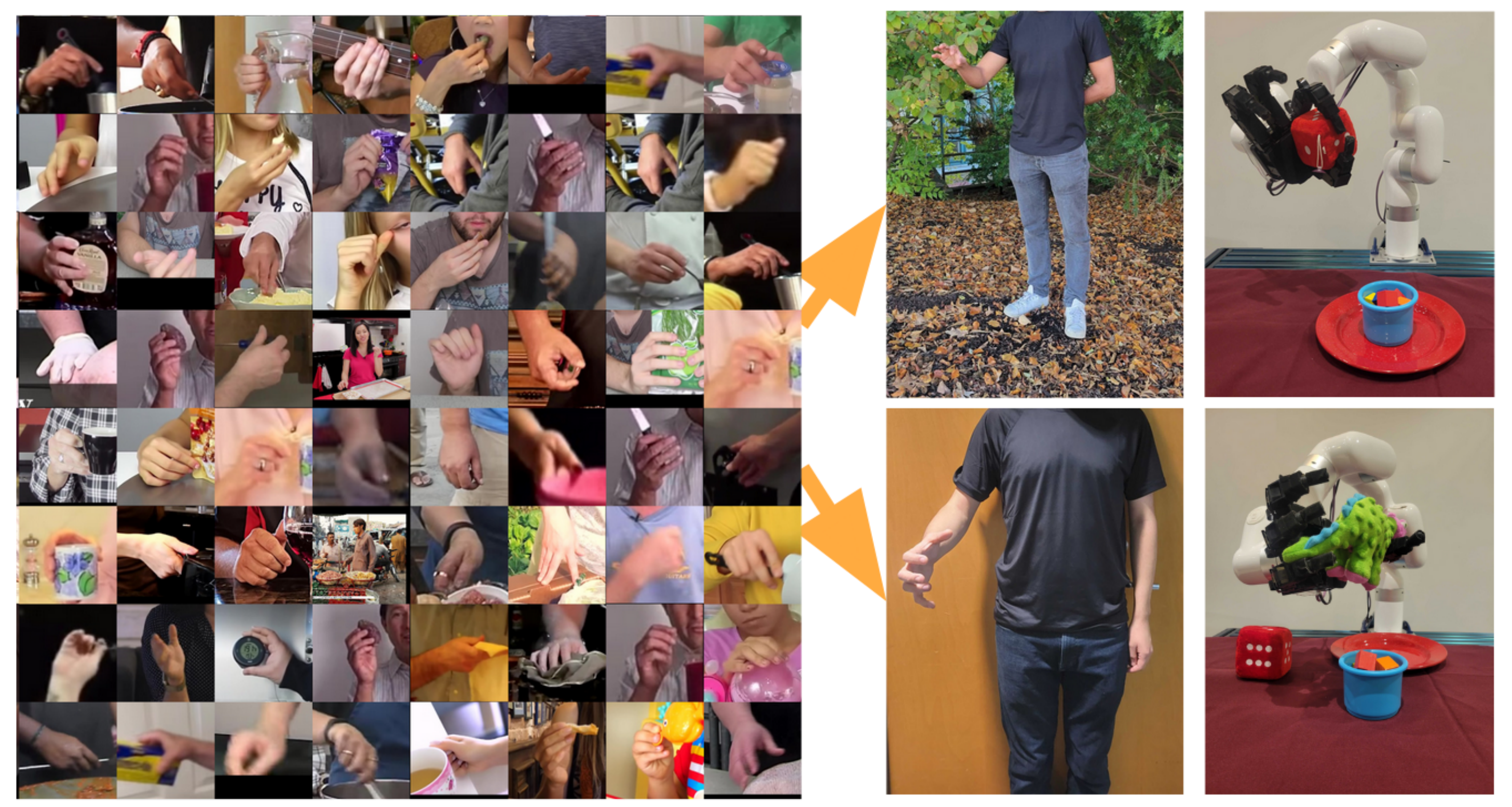}
 \vspace{-0.2in}
\caption{\small{Our system leverages passive data from the internet to enable robotic real-time imitation in-the-wild. This low-cost system does not require any special gloves, mocap markers or even camera calibration and works from a single RGB camera.}}
\vspace{-0.1in}
 \label{fig:teaser}
\end{figure}

The challenge of overcoming these shortcomings grows exponentially with the complexity of the robot to be controlled; multi-fingered hands are far more difficult to teleoperate than two-finger grippers.  Despite recent advancements, building an easy-to-use, performant and low-cost teleoperation system for high-dimensional dexterous manipulation has remained elusive.  Handa \textit{et al.} recently proposed DexPilot ~\cite{dex}, a low-cost system for vision-based teleoperation that is free of markers or hand-held devices.  It lowers the cost and usability barrier, but relies on a custom setup with multiple calibrated depth cameras, and uses neural networks trained on images collected in this controlled environment, which limits its use to a specific lab setting.

\textbf{The objective of this paper is to enable teleoperation of a dexterous robotic hand, \textit{in the wild}. This means our system should be low-cost, work for any untrained operator, in any environment, with only a single uncalibrated color camera.} One should be able to simply look into a monocular camera of their phone or tablet and control the robot without relying on any bulky motion capture or multi-camera rigs for accurate 3D estimation. We call our system \textit{Robotic Telekinesis}, as it provides a human the ability to control a dexterous robot from a distance without any physical interaction.

Unfortunately, building such a system poses a chicken-and-egg problem: to train a teleoperation system that can work in the wild, we need a rich and diverse dataset of paired human-robot pose correspondences, but to collect this kind of data, we need an in-the-wild teleoperation system.  However, while we lack paired human-robot data, there is no shortage of rich human data, and \textbf{our key insight is to leverage a massive unlabeled corpus of internet human videos} at training time.  These videos capture many different people from different viewpoints doing different tasks in several environments, ensuring generalization by design.

We propose a method that conquers the human-to-robot problem using two subsystems.  The first subsystem uses powerful computer vision algorithms trained to estimate 3D human poses from 2D images, and the second subsystem uses a novel motion-retargeting algorithm to generate a physically plausible robot hand-arm action that is consistent with a given human pose.  During training, our method only uses \textit{passive data} readily available online and does not require any \textit{active} fine-tuning on our robot in our lab setup.

Our system is low-cost, glove-free, and marker-free, and it requires only a single uncalibrated color camera with which to view the operator. It allows any operator to control a four-finger 16 Degree-of-Freedom (DoF) Allegro hand, mounted on a robotic arm, simply by moving their own hand and arm, as illustrated in Figure~\ref{fig:teaser}. We demonstrate the usability and versatility of our system on ten challenging dexterous manipulation tasks.  We further demonstrate the generality and robustness of our system by performing a systematic study across ten previously untrained human operators.

\begin{figure}[t]
 \centering
 \includegraphics[width=\linewidth]{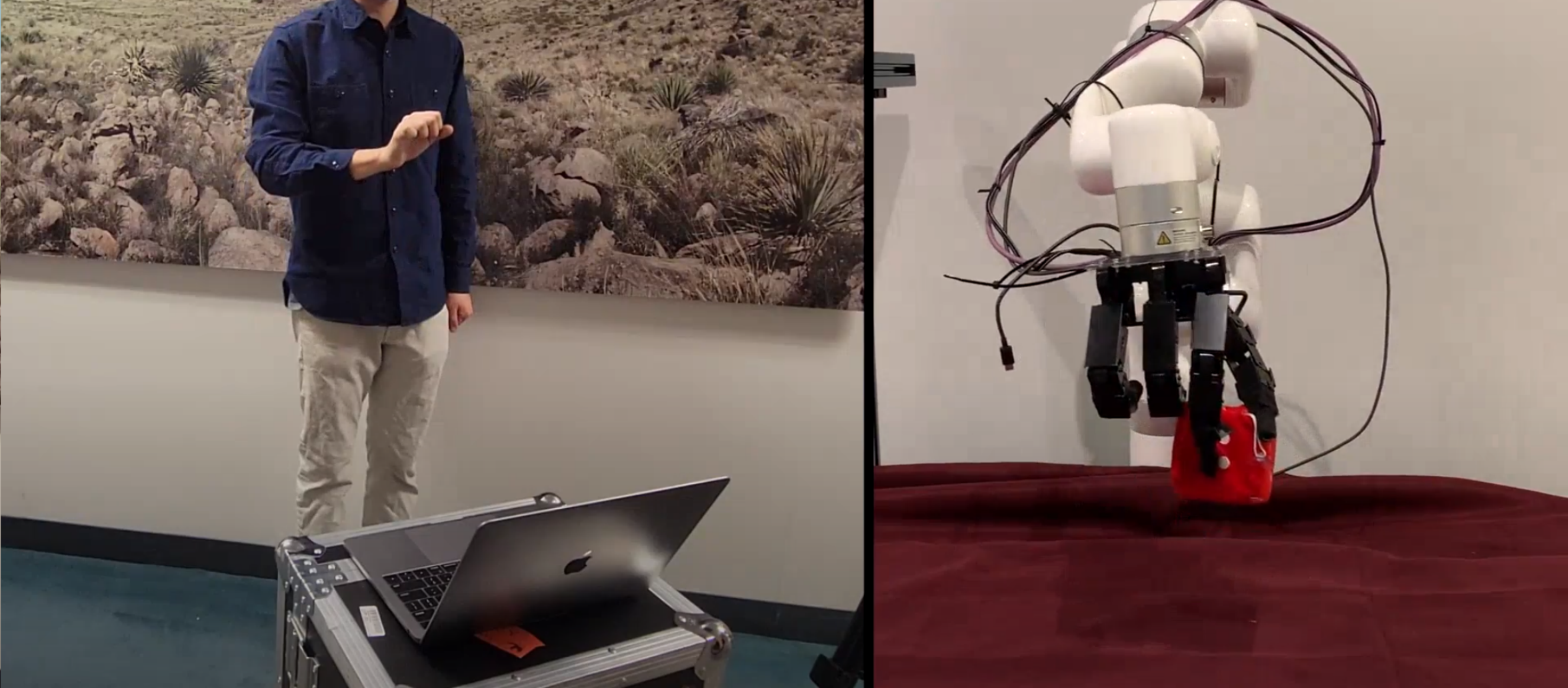}
\caption{\small{An operator completing a dice pickup task while watching the robot through a video conference. Video demos are at \href{https://robotic-telekinesis.github.io/}{https://robotic-telekinesis.github.io/}.}}
 \label{fig:teleoperation_figure}
 \vspace{-0.15in}
\end{figure}

\section{Related Work}
\label{sec:related_works}
{The first section reviews relevant research in 3D human pose estimation and the second section discusses related work in kinematic motion retargeting, with a particular focus on cross-embodiment retargeting and teleoperation.}

\textbf{Understanding Human Hands and Bodies}  Modeling human bodies and estimating their poses are widely studied problems, with applications in graphics, virtual reality and robotics.  The recent research most relevant to our work can roughly be divided into four sub-areas.  (1) \textit{Hand and body modeling.}  MANO \cite{MANO:SIGGRAPHASIA:2017} is a low-dimensional parametric model of a human hand, and SMPL \cite{loper2015smpl} is an analogous model for the human body.  SMPL-X \cite{SMPL-X:2019} is a single unified model of both the body and hands.  (2) \textit{Monocular human hand and body pose estimation}.  Recent works in human pose estimation typically estimate 2D quantities like bounding boxes \cite{100doh} or skeletons \cite{cao2019openpose}, or perform a full 3D reconstruction \cite{wang2020rgb2hands, hmr, feng2021collaborative}.  Rong \textit{et al.} \cite{FrankMocap_2021_ICCV} propose a method for integrated 3D reconstruction of human hands and bodies. (3) \textit{Dataset curation}: The advances in human pose estimation crucially rely on large datasets of human hand and body poses.  FreiHand \cite{Freihand2019}, Human3.6M \cite{ionescu2013human3} and the CMU Mocap Database \cite{cmu_mocap} are examples of densely-annotated datasets in clean indoor settings.  On the other end of the spectrum, the 100 Days of Hands \cite{100doh} and Epic Kitchens \cite{EPICKITCHENS} datasets are massive collections of raw videos that span a rich and diverse set of hand poses and motions but don't contain pose annotations.   (4) \textit{Understanding human hand function}.  Brahmbhatt \textit{et al.} \cite{Brahmbhatt_2019_CVPR} use thermal imaging to capture impact heatmaps that reveal patterns in the ways human hands interact with everyday objects.  Taheri \textit{et al.} \cite{taheri2020grab} study the problem of how human hands and bodies behave while grasping and manipulating objects, and propose a method to generate plausible human grasps for novel objects.  Hasson \textit{et al.} \cite{hasson19_obman} and Cao \textit{et al.} \cite{cao2021reconstructing} propose methods for joint hand and object reconstruction, and Hampali {et al.} \cite{hampali2020honnotate} present a dataset of hand-object interactions with 3D annotations.
Liu \textit{et al.} \cite{Liu2014ATO} builds a taxonomy of human grasps to understand the cognitive patterns of human hand behavior~\cite{light2002establishing}.

\textbf{Kinematic Retargeting and Visual Teleoperation}
Human pose estimation only solves half of the visual teleoperation problem.  Mapping human poses to robot poses is itself a difficult challenge, because humans and robots have very different kinematic structures.  Li \textit{et al} \cite{li2019vision} train a deep network to map human hand depth images to joint angles in the robotic Shadow Hand, and Antotsiou \textit{et al} \cite{antotsiou2018task} combine inverse kinematics and Particle Swarm Optimization to retarget human hand poses to a high-dimensional robot hand model.  Our system follows the method of DexPilot \cite{dex}, which minimizes a cost function that captures the functional similarity between a human and a robot hand.

The general problem of kinematically retargeting motion in one morphology into another is also studied outside of robotic manipulation.  Villegas \textit{et al.} \cite{neural_kin} propose a cycle consistency objective to transform motion between animated humanoid characters of different body shapes.  Peng \textit{et al} \cite{peng2020learning} use an approach based on keypoint matching to learn robotic locomotion behaviors from demonstrations of walking dogs.  Zakka \textit{et al.} \cite{zakka2021xirl} learn a visual reward function that allows reinforcement learning agents to learn from demonstrators with different embodiments.

\begin{figure*}[t]
 \centering
 \includegraphics[width=\linewidth]{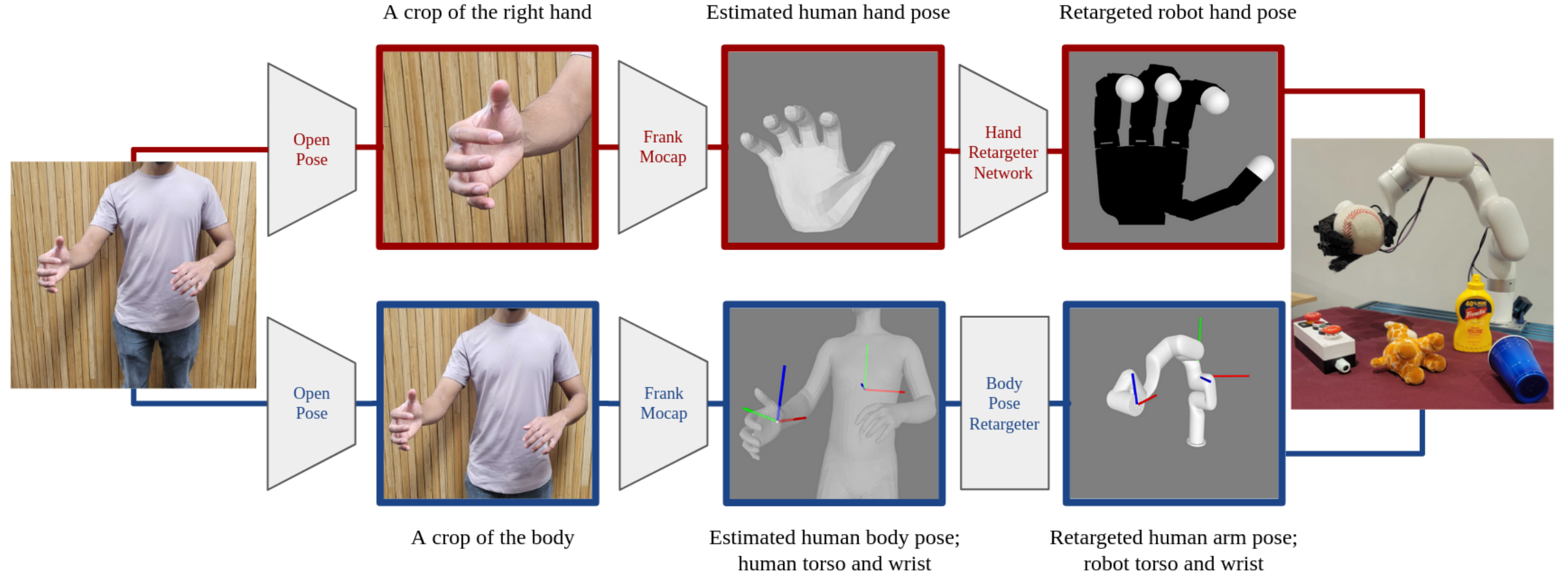}
 \vspace{-0.25in}
 \caption{\small A graphical description of our visual teleoperation pipeline.  First, a color camera captures an image of the operator.  {Top:} to command the robot hand, a crop of the operator's hand is passed to a hand pose estimator,  and the hand retargeting network maps the estimated human hand pose to a robot hand pose.  {Bottom:} to command the robot arm, a crop of the operator's body is passed to a body pose estimator and cross-body correspondences are used to determine the desired pose of the robot's end-effector from the estimated human body pose.  Finally, commands are sent to both the robot hand and arm.}
 \vspace{-0.1in}
 \label{fig:method}
\end{figure*}

\section{Robotic Telekinesis}
\label{sec:method}

{\textit{Robotic Telekinesis} is a system for real-time, remote visual teleoperation of a dexterous robotic hand and arm.  A human demonstrates tasks to the robot just by moving their hand, and the robot mimics the actions instantaneously.  Our system consists of an xArm6 robot arm, a 16-DoF Allegro robot hand, and a single uncalibrated RGB camera capturing a stream of images of the human operator.}  {The operator must be in the field of view of the camera and must be able to see the robot to guide it, either in real life or through a video conference feed.}  {Each image is \textit{retargeted} into two commands which place the robot hand and arm in poses that match the hand-arm poses of the human operator in real time.  Figure ~\ref{fig:teleoperation_figure} illustrates an operator solving a grasping task while monitoring the robot through a video conference feed.}

The problem of remote teleoperation from a single camera is severely under-constrained for two reasons. One reason is that the input images are in 2D while the robot is controlled in 3D: mapping from 2D to 3D is an ill-defined problem. {While this issue could be addressed with a multi-camera setup, in this work our goal is to build a system usable with any one uncalibrated camera, from a cell-phone camera to a cheap webcam. The use of a single color-only camera leads to certain failure modes, typically related to inter-hand occlusions and ambiguous depth perception, but these are issues we attempt to mitigate using our neural network based retargeter.  }

The second reason is the ambiguity caused by the differences in morphology, shape and functionality, between human hands and robot hands.  To address both of these problems, we rely on deep neural networks to learn priors from passively-collected internet-scale human datasets to enable powerful human pose estimation and human-to-robot transfer.

In the rest of this section, we describe our visual teleoperation pipeline. As shown in Figure~\ref{fig:method}, we group the pipeline into two branches: one branch for hand retargeting and the other one for arm retargeting.

\subsection{\textbf{Hand Teleoperation: Human Hand to Robot Hand Pose}}
The problem of retargeting 2D human images to robot hand control commands is broken into two sub-problems.  The first is to estimate the 3D pose of the human hand from a 2D image, and the second is to map the extracted 3D human hand parameters to robot joint control commands. We discuss each of these two sub-problems below.

\vspace{2mm}\subsubsection{\textbf{2D Hand Image to 3D Human Hand Pose}}
\label{3d-estimation}

{The first step in hand retargeting is to detect the operator's hand in a 2D image and infer its 3D pose.  To this end, we exploit recent advances in computer vision.  We rely on large paired 2D-3D datasets, and high-quality models that leverage this data to produce physically plausible 3D human pose estimates from  2D images.}

Our method first computes a ``crop" around the operator's hand, based on a bounding box computed using an off-the-shelf detector derived from OpenPose \cite{cao2019openpose}.  The resulting image crop goes to a pose estimator from FrankMocap \cite{FrankMocap_2021_ICCV} to obtain hand shape and pose parameters of a 3D MANO model \cite{MANO:SIGGRAPHASIA:2017} of the operator's right hand. 
See the ``OpenPose" and ``FrankMocap" modules in the top row of Figure \ref{fig:method} for a graphical depiction of this phase of the pipeline, and see the appendix for further implementation details.
  
We emphasize that our human hand pose estimation module works for \textit{any human operator}, with \textit{any camera} in \textit{any environment}, without any further fine-tuning.  This strong generalization is due to the diversity of millions of images on which the neural network and pose estimators are trained.

\vspace{2mm}\subsubsection{\textbf{3D Human Hand to Robot Hand Pose}}
\label{human2robot}

{Next, estimated 3D human hand poses are retargeted to the 16 Allegro hand joint angles to place it in an analogous hand pose (see the third panel on the top branch of Figure \ref{fig:method}).}  This has three challenges:
\begin{itemize}
    \item \underline{Underconstrained}: The Allegro hand and the human hand have many DOF and very different embodiments: they differ greatly in shape, size and joint structure.
    \item \underline{{Generality}}: Our retargeter must work for \textit{any human operator} trying to perform \textit{any kind of task} in \textit{any environment}. 
    \item \underline{Efficiency}: {We require a real-time solution ($>$15 Hz).} 
\end{itemize}

A natural way to address these three challenges would be to train a {supervised learning model} on a diverse dataset of paired human-robot hand pose examples. However, collecting this large scale dataset would be prohibitively expensive. Instead, we train a deep human-to-robot hand retargeter network that uses human data alone.

\begin{figure}[t]
 \centering
 \includegraphics[scale=0.3]{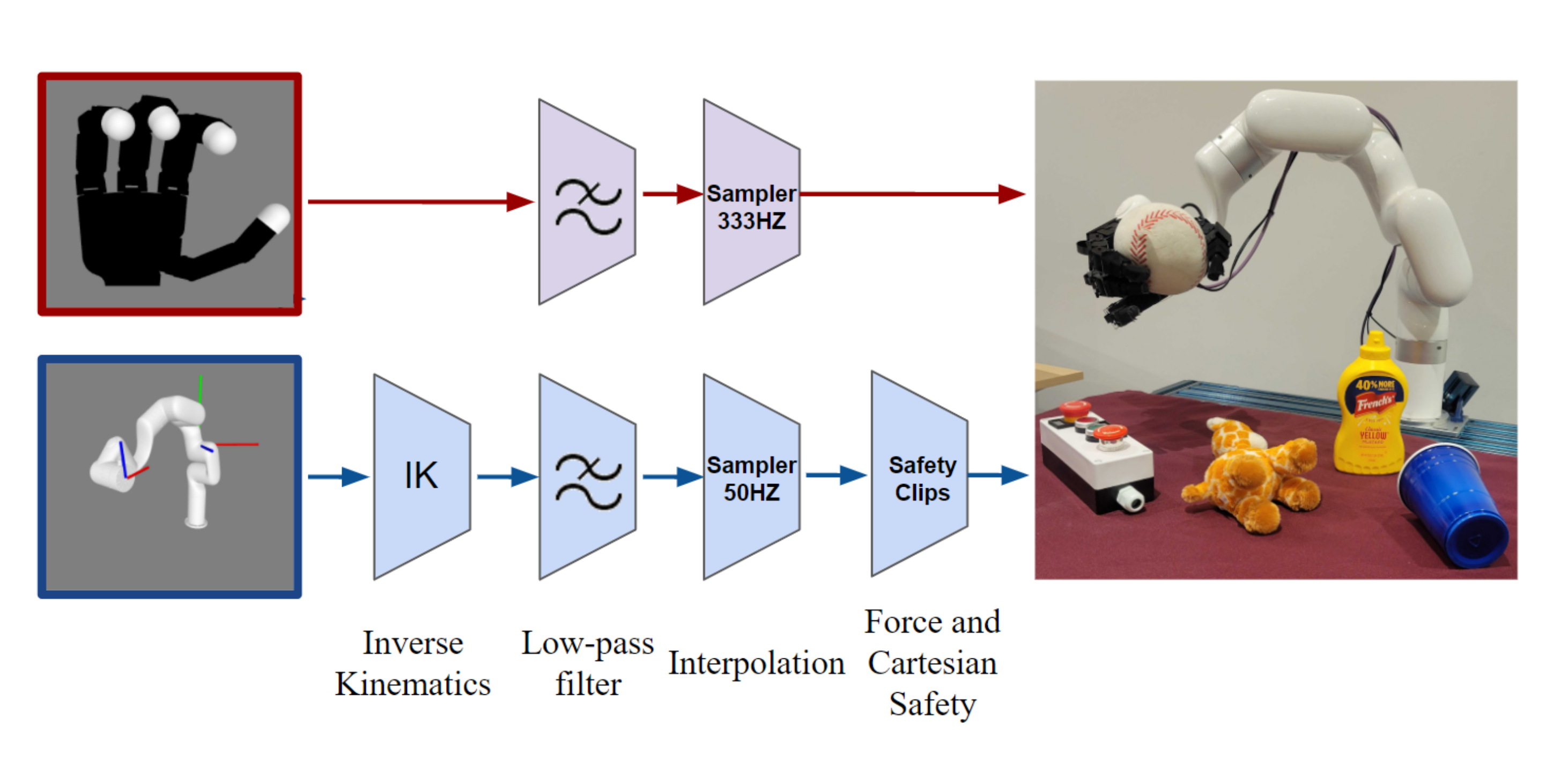}
 \vspace{-0.2in}
\caption{\small \textbf{Control Pipeline}. Description of our control stack.  {Raw target poses are received from the visual retargeting modules, then} Inverse kinematics, low-pass filtering, sampling, and safety clipping are performed. {The smoothed commands are sent to the robot.}}
 \vspace{-0.08in}
 \label{fig:control_function}
\end{figure}

\vspace{2mm}\noindent\textbf{Dataset of YouTube Videos of Human Interaction$\quad$} 
We leverage a massive internet-scale dataset of human hand images and videos.  We gather about 20 million images from the Epic Kitchens \cite{EPICKITCHENS} dataset, which captures ego-centric videos of humans performing daily household tasks, and the 100 Days of Hands \cite{100doh} dataset, which is a collection of YouTube videos.  We run the hand pose estimator from \cite{FrankMocap_2021_ICCV} (the same one we use at deployment time in our pipeline) to estimate human hand poses for each image frame in these videos.  We augment this massive noisy dataset of estimated human hand poses with the small and clean FreiHand dataset ~\cite{Freihand2019}, which contains ground-truth human hand poses for a diverse collection of realistic hand configurations. 

\vspace{2mm}\noindent\textbf{A Lack of Paired Data$\quad$}
Our dataset contains millions of human hand poses, but no ground-truth target robot poses to regress onto.  In most neural network {regimes}, at training time, the network has access to \textit{paired} examples $(x \in \mathcal{X}, y \in \mathcal{Y})$, where $\mathcal{X}$ is the source domain, and $\mathcal{Y}$ is the target domain.  In our case, the source domain $\mathcal{X}$ is the set of all human hand poses and the target domain $\mathcal{Y}$ is the set of all robot hand poses, \textit{but we only have training data from the source domain.} Hence, we can not perform a direct regression. 

\vspace{2mm}\noindent\textbf{Energy Function Formulation$\quad$}
Instead, we formulate the retargeting problem using a feasibility objective.  We posit that the optimal corresponding robot hand pose is the one that best mimics the \textit{functional intent} of the human. In order for the robot hand to effectively mimic human actions, the relative positions between the robot's fingertips should match those of the human's.  Following~\cite{dex}, we define a set of five hand keypoints (4 fingertips (no pinky) and a palm), and ten keyvectors which connect all pairs of keypoints. 

These keyvectors are used in an \textit{energy function} that captures \textit{dis}similarity between a human hand pose (parameterized by the MANO model parameters $(\beta_{h}, \theta_{h})$) and an Allegro hand pose (parameterized by the joint angles $q_{a}$).  First, for each $i \in \{1, \dots, 10\}$, the $i$-th keyvector is computed on the human hand (call it $\mathbf{v_i^h}$) and the Allegro hand (call it $\mathbf{v_i^a}$).  Then, each Allegro hand keyvector $\mathbf{v_i^{a}}$ is scaled by a constant $c_i$.  The $i$-th term in the energy function is then the Euclidean difference between $\mathbf{v_i^h}$ and $c_i \cdot \mathbf{v_i^a}$:
\begin{equation}
    E( \ (\beta_{h}, \theta_{h}), \  q_{a} \ ) = \sum_{i=1}^{10} || \mathbf{v_i^h }- (c_i \cdot \mathbf{v_i^a}) ||_2 ^2
    \label{eq:energy}
\end{equation}
where the scaling constants $\{c_i\}$ are hyperparameters.  Critically, this energy function is a fully differentiable function of the Allegro joint angles (because of the differentiability of the forward kinematics operation), which allows us to train the hand retargeter network $f$ via gradient descent, using the energy function as a loss function.

 This energy optimization formulation is different than the prototypical IK problem, which solves for joint angles that achieve target fingertip poses relative to a fixed base. Our end-effector constraints are all relative to each other, which makes it difficult to adopt open-source IK solvers such as \cite{carpentier2019pinocchio}.

\begin{figure}[t]
 \vspace{-0.1in}
 \centering
 \includegraphics[scale=0.45]{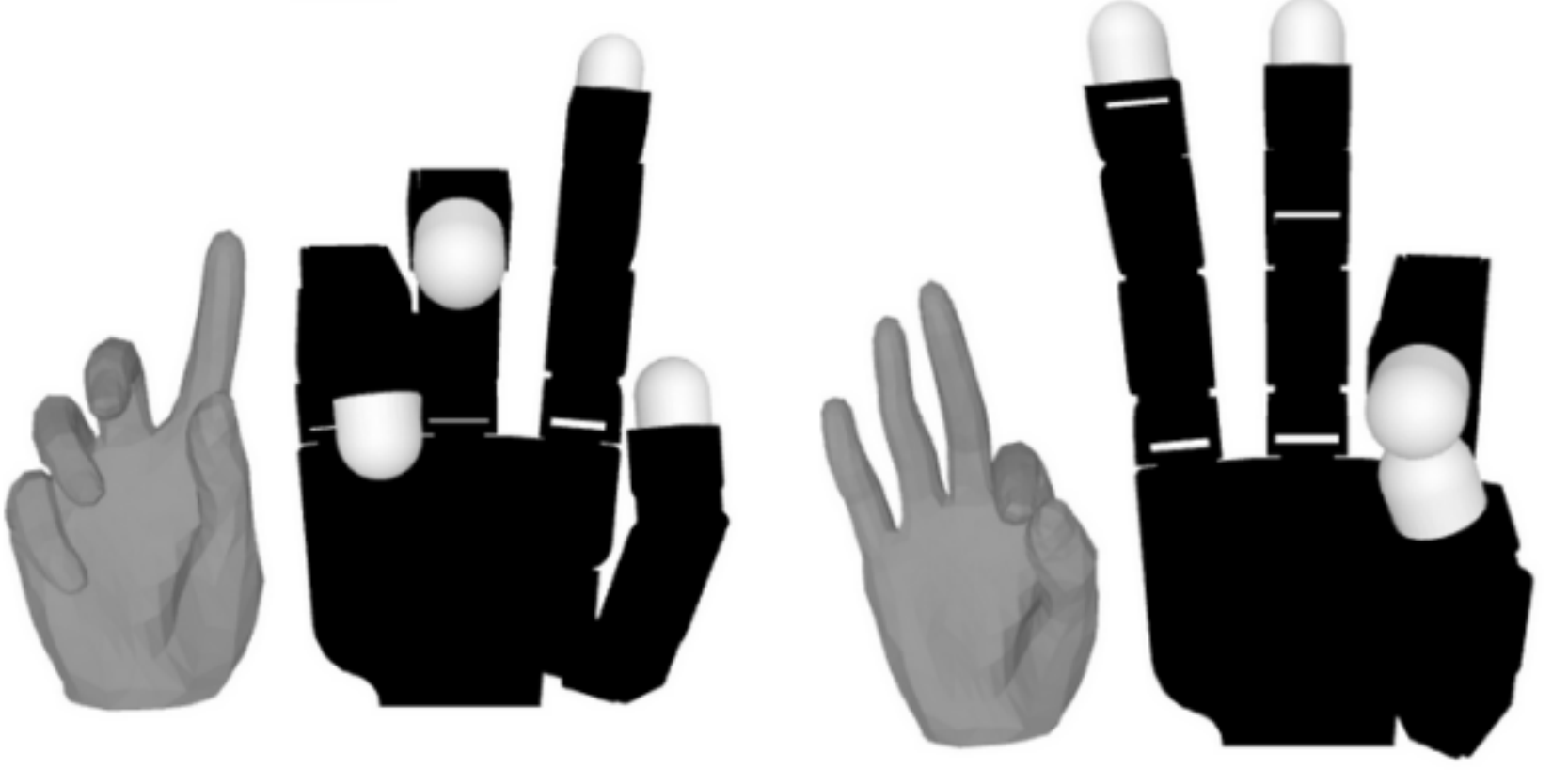}
 \vspace{-0.1in}
\caption{\small \textbf{Human-to-robot Translations}.  The inputs and outputs of our hand retargeting network.  Each of the pairs depicts a human hand pose, and the retargeted Allegro hand pose.}
 \vspace{-0.15in}
 \label{fig:energy_function}
\end{figure}

\vspace{2mm}\noindent\textbf{Retargeter Network$\quad$}
Our hand retargeter network is a Multi-Layer Perceptron (MLP), $f(.)$, with two hidden layers.  It takes as input a human hand pose (a vector $x \in \mathbb{R}^{55}$ that denotes the MANO hand shape and pose parameters) and outputs a vector of Allegro joint angles $y \in \mathbb{R}^{16}$.
Since there is no paired labels available for human to robot hand, our network is trained to minimize the energy function $E(x, y)$ in Equation~\ref{eq:energy} that captures the dissimilarity between the input 3D human hand pose $x$ and the network's predicted robot hand pose $y$. Per convention, a high energy means the two poses are highly \textit{dis}similar. Formally, the neural network optimizes the following objective:
\begin{equation}
    \underset{f}{\mathrm{argmin}} \  \mathbb{E}_{x \in \mathcal{X}} \ \bigg[ E(x, f(x)) \bigg],
    \label{eq:network-training}
\end{equation}

At inference time, we simply pass the estimated hand shape/pose vector to the network, which directly outputs Allegro joint angles that we can command to the robot.  A key benefit of using a neural network is speed: the network's forward pass takes about 3ms (333Hz) -- this is critical for smooth real-time teleoperation.

\vspace{2mm}\subsubsection{\textbf{Collision Avoidance via Adversarial Training}}
\label{sec:collision}
Using a neural network to perform human-to-robot hand retargeting has another subtler advantage over an online optimization approach: {we can augment the energy function with terms that are slow to compute.} When training a neural network, we can run expensive operations in order to compute the loss at each iteration.  This allows us to use any energy function, as long as it is differentiable.  We simply absorb the computation cost during training instead of incurring it at deployment time.

\textit{We exploit this idea to address the problem of self-collisions.}  Minimizing the keyvector-similarity energy function described above can sometimes yield robot hand joint configurations which orient the hand such that fingers collide with each other or with the palm.  It is difficult to add a term to the energy function that penalizes such configurations, since ``self-collision-ness" is not a differentiable function of the robot’s joint angles.

To address this, we first train a classifier that takes in an Allegro joint angle vector, and tries to classify whether or not the joint configuration yields a self-collision.  This classifier is an MLP, and we generate its training data programatically by repeatedly sampling a joint angle vector within the legal joint limits, and querying a (non-differentiable) self-collision checker to generate a ground-truth binary self-collision label.

\begin{figure}[t]
 \centering
 \includegraphics[scale=0.3]{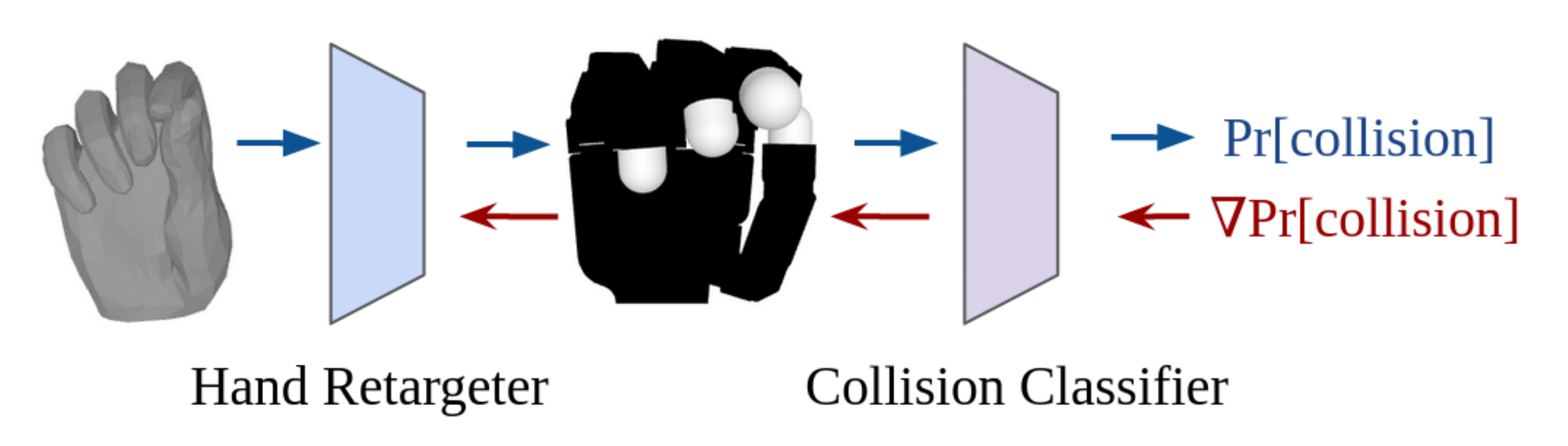}
 \vspace{-0.05in}
\caption{\small A trained self-collision classifier is used as {an adversary} that penalizes self-colliding joint configurations. The blue arrows denote the forward pass, and the red arrows denote the flow of gradients during the backward pass.}
\vspace{-0.1in}
 \label{fig:discriminator}
\end{figure}

Once our self-collision classifier is trained, we use it as a “discriminator” to train our retargeter network.  At every training iteration, we pass the retargeter's predicted robot joint angle vectors to the self-collision classifier.  Intuitively, we want the predicted self-collision score to be as low as possible, and therefore we use it as a term in the loss function for the retargeter network.  \textit{The gradient of the self-collision score from the collision network is backpropagated through the self-collision classifier, and used to update the weights of the retargeter network}, as shown in Figure \ref{fig:discriminator}.  This leads the retargeter network to avoid outputting Allegro joint angle configurations that the self-collision classifier believes to be illegal.  Our retargeter network and self-collision classifier are akin to the generator and discriminator respectively in a Generative Adversarial Network (GAN), though in our case, we pretrain and freeze the self-collision classifier so we don’t suffer the instability of jointly optimizing a discriminator and a generator, notorious in GAN training.

\subsection{\textbf{Arm Teleoperation: Human Body to Robot Arm Poses}}

A hand that can flex its fingers but does not have the mobility of an arm will not be able to solve many useful tasks.  The second branch of our retargeting pipeline therefore focuses on computing the correct pose for the robot arm from images of the human operator.

\begin{figure*}[t]
 \centering
 \includegraphics[width=\linewidth]{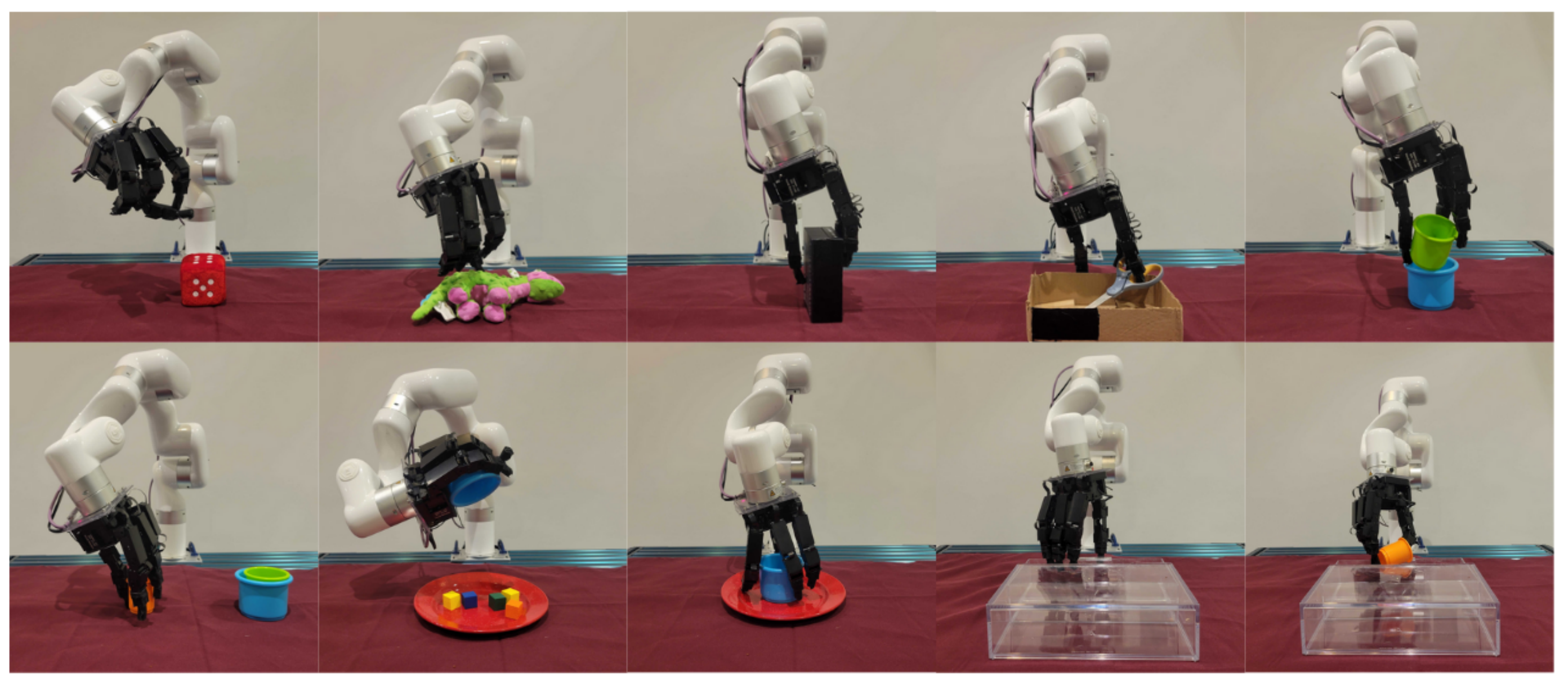}
 \vspace{-0.2in}
 \caption{\small {Ten different teleoperation tasks.} {Top row, left to right:} Pickup Dice Toy, Pickup Dinosaur   Doll, Box Rotation, Scissor Pickup, Cup Stack.  {Bottom row, left to right:} two cup stacking, pouring cubes onto plate, cup into plate, open drawer and open drawer and pickup cup. Please see videos {at \href{https://robotic-telekinesis.github.io/}{https://robotic-telekinesis.github.io/}.}}
 \label{fig:tasks}
\end{figure*}

\begin{table*}[t]
\begin{center}
\resizebox{\linewidth}{!}{%
\begin{tabular}{lcc| llllr} 
 \toprule
  & \multicolumn{2}{c}{Success (rate)} & \multicolumn{2}{c}{Completion Time (sec)} & \\
 Task & Ours & \shortbaseline & Ours & \shortbaseline & Description\\ %
 \midrule
 Pickup Dice Toy & \textbf{0.9} & 0.7 & \textbf{8.6 (2.65)} & 13.5 (5.47) & Pickup Plush dice from table.  \\ %
 Pickup Dinosaur Doll & \textbf{0.9} & 0.6 & \textbf{8.2 (3.49)} & 11.00 (3.95) &  Pickup Plush dinosaur from table.\\
 Box Rotation & \textbf{0.6} & 0.3 & 37.2 (12.6)  & \textbf{16.33 (10.69)} & Rotate box 90 degrees onto the smaller side\\
 Scissor Pickup & \textbf{0.7} & 0.5 & 28.6 (9.4) & \textbf{27.66 (11.09)} & Remove Scissors from the box using fingers\\
 Cup Stack & 0.6 & \textbf{0.7} & \textbf{21.5 (7.6)} & 22.85 (16.57) &  Smaller cup must be placed inside the large cup.\\
 Two Cup Stacking & \textbf{0.3} & 0.1 & \textbf{27.3 (11.0)} & 45.00 (0.0) &  Small cup placed into medium cup into large cup.\\ %
 Pouring Cubes onto Plate & \textbf{0.7} & 0.5 &  36.80 (17.7) & \textbf{13.8 (4.02)} & Five cubes in a cup must be poured onto a plate.\\ %
 Cup Into Plate & \textbf{0.8} & 0.7 & \textbf{10.6 (4.4)} & 13.71 (5.44) & Place cup on the plate.\\ %
 Open Drawer & \textbf{0.9} & \textbf{0.9} & 23.6 (12.3) & \textbf{14.88 (4.40)} & Open clear drawer.\\ %
 Open Drawer and Pickup Cup & 0.6 & \textbf{0.7} &  33.7 (8.1) & \textbf{28.14 (11.48)} & Open clear drawer and pickup cup inside.\\ %
  \bottomrule
\end{tabular}}
\caption{\small
{Success rate and completion time (mean and standard deviation) of a trained operator completing a variety of tasks using two different methods. The \texttt{\baseline} baseline is nearly identical to our system, but uses online gradient descent for hand pose retargeting (inspired by DexPilot~\cite{dex}).  Our system, which uses a neural network retargeter, outperforms the baseline in 7 out of 10 tasks.}}
\label{tab:accuracy}
\end{center}
\end{table*}

 \begin{figure*}[t]
 \centering
 \includegraphics[scale=0.6]{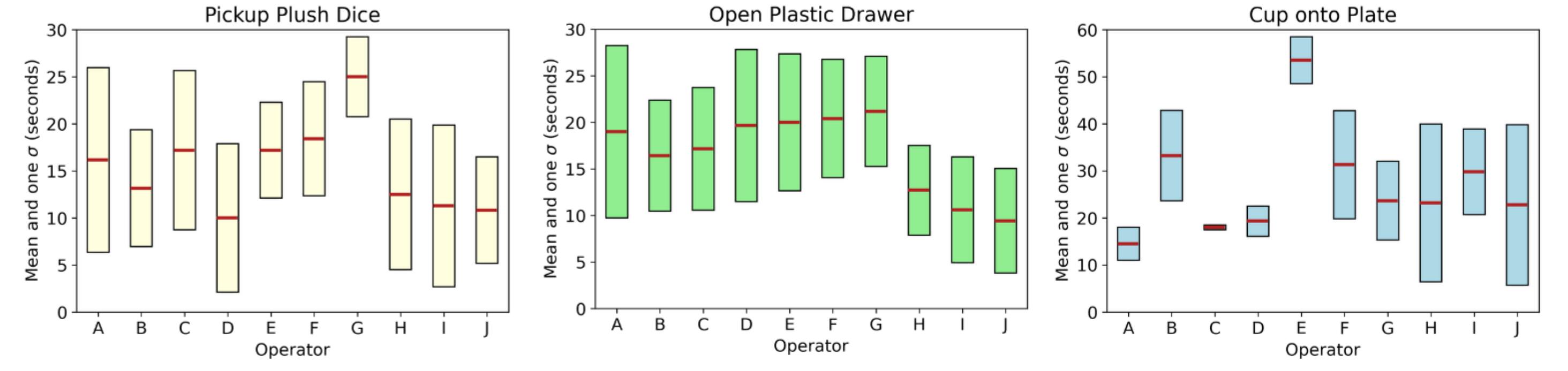}
 \vspace{-0.1in}
 \caption{\small Ten {novice} operators were asked to complete tasks: (1) picking up a plush dice, (2) opening a plastic drawer, and (3) placing a cup onto a plate. {For each task, the mean and standard deviation completion times were computed over seven trials.}}
 \label{fig:plots}
 \end{figure*}

At each timestep, we first compute a crop of the operator's body using a bounding box detector derived from OpenPose \cite{cao2019openpose}, then pass the crop to the body pose estimator from FrankMocap \cite{FrankMocap_2021_ICCV}.  We model the human body using the parametric SMPL-X model, and the body pose estimator predicts the 3D positions of the joints on the human kinematic chain.  

Because we aim to build a system that operates from a single {``floating"} color camera, there are two main problems that arise.  (1) Without a depth sensor or camera intrinsics, we cannot accurately estimate how far from the camera the human's {hand} is.  (2) Without camera-to-robot calibration, there is no known transformation between the camera, robot and human.  ({With a calibrated depth camera}, we would simply mount a camera with a fixed known transformation relative to the robot’s base frame, localize the position and orientation of the operator’s wrist in the 3D camera coordinate frame, and use the known camera extrinsics to determine how the robot’s wrist should be positioned and oriented.)

Instead, we \textit{estimate the relative transformation between the human wrist and an anchor point on the human's body}.  We define the human's torso as the origin, and manually choose a suitable point to serve as the “robot’s torso”.  We posit that the relative transformation between the human's right hand wrist and torso should be the same as the relative transformation between the robot's wrist link and the robot's torso.  By traversing the kinematic chain from the torso joint to the right hand wrist joint, we compute the relative position and orientation between the human's right hand wrist and torso (see Figure~\ref{fig:method}). The bottom row in Figure~\ref{fig:method} depicts this pipeline visually. This simple correspondence trick works surprisingly well in practice and provides a natural user experience for even a moving operator.  

To handle minor errors in human body pose estimation and ensure smooth motion, we reject outliers, and apply a low-pass filter on the stream of estimated wrist poses. We then use an IK solver \cite{SDLS} to compute arm joint angles that place the robot’s end-effector (i.e. ``wrist") at the correct relative transformation relative to the "robot torso" coordinate frame. See the bottom row of figure~\ref{fig:control_function} for a depiction of our control stack, and see the appendix for further details about the arm retargeting modules.

\section{Experimental Results}
\label{sec:experiment}
We evaluate the strengths and limitations of our system through experiments on a diverse suite of dexterous manipulation tasks with an expert operator.  We also demonstrate the usability and robustness of the system through a smaller set of tasks on a group of ten previously untrained operators. Videos can be found at \href{https://robotic-telekinesis.github.io/}{https://robotic-telekinesis.github.io/}.

\vspace{2mm}\noindent
{\textbf{Baseline$\quad$}  Our hand retargeter neural network is compared to an \textit{online optimization} procedure that runs online gradient descent to minimize the energy function between the human and robot hand.  We call this baseline \texttt{\baseline}: the use of online optimization for retargeting is modeled after DexPilot~\cite{dex}, but the overall system (including the single-camera setup) is held constant between the baseline and our method.  At each timestep, given an estimated human hand pose $x$, a solver iteratively searches for the robot pose $y^*$ that minimizes the energy (cost) function $\mathcal{L}$ with respect to $x$, i.e. 
\begin{equation}
    y^* = \underset{y}{\mathrm{argmin}} \ \mathcal{L}(x, y).
\end{equation}
The code for DexPilot's~\cite{dex} kinematic retargeting module is not available, so we implement their online optimization solver using the Jax GPU-accelerated auto-differentiation engine~\cite{jax2018github}.

We do not compare our system to the full DexPilot system.  DexPilot is designed for use in a specific multi-camera rig, but our system is designed to run anywhere.  The \texttt{\baseline}  baseline is meant to enable analysis of the tradeoffs between online optimization and neural networks for kinematic retargeting, within a single-camera setup.  It uses the retargeting module from DexPilot~\cite{dex}, but is otherwise identical to our system.}

\subsection{Success Rate: Trained Operator Study}
A trained operator attempted a diverse set of tasks to test the capabilities of our system and the \texttt{\baseline} baseline. These tasks are shown in Figure~\ref{fig:tasks}.  They span a diverse spectrum of arm and hand motions and involved interacting with a variety of different objects.  Each of the ten tasks was run for ten trials with a timeout period of one minute.  This rigorously tested the system's capabilities and limitations. These tasks are described in Table~\ref{tab:accuracy}.  {The operator achieved good success on all tasks -- our system outperformed the baseline on 7 out of 10 tasks, and performed similarly on the other 3 tasks.}  Grasping plush objects proved easy as these grasps do not require much precision, but we observed that fine-grained grasps of smaller, more slippery objects like plastic cups occasionally proved difficult.  See videos of the trained operator completing these tasks at: \href{https://robotic-telekinesis.github.io/}{https://robotic-telekinesis.github.io/}. 
During experiments, the expert found that our system was easier to use and performed better than \texttt{\baseline}.  The online gradient descent solver in the baseline occasionally stayed stuck in local minima because it would use the previous pose as a seed.  This meant that the hand would often output unnatural poses with the fingers digging into the palm, an issue that {the authors of} DexPilot also noted. Our method, because it was trained on YouTube data, learned to always output natural hand poses which was useful for operators to use.  Since it is not seeded, our method did not get stuck in minima.  This data also masked the ambiguities and errors from our single camera constrained setup.  Our method also produced occasional errors on uncommon hand poses unseen in the training set, but these one-off errors did not propagate forward through time.  Additionally, the baseline ran at a slower rate and felt delayed to the operator's movements as benchmarked in table \ref{tab:timing_comparison}. This was jarring and hard to compensate for when trying to complete dexterous tasks.  Our system maintained fluidity and felt very responsive when opening and closing the hand. 

\begin{table}[t]
\centering
\resizebox{0.8\columnwidth}{!}{%
\begin{tabularx}{\columnwidth}{lc}
 \toprule
 Pipeline Stage & Ours (Hz)\\ %
 \midrule
 Open Pose Body \scriptsize{(input from camera)} & 29\\ %
 Open Pose Hand \scriptsize{(input from camera)} & 29\\ %
 Frank Mocap Body & 16\\ %
 Frank Mocap Hand & 27 \\ %
 Body Pose Retargeter \scriptsize{(output to robot)}  & 16\\ %
 Hand Retargeter \scriptsize{(output to robot)}  & 24\\ %
  \bottomrule
\end{tabularx}}
\caption{\small Runtime of each stage of our pipeline.  Our hand retargeter NN runs at 24 Hz (the online gradient-descent baseline runs at 10Hz).  Both systems use an AMD 3960x CPU and two 3080 Ti GPU's.}
\label{tab:timing_comparison}
\end{table}

\subsection{Usability: Human-Subject Study}
To test usability and generality, we conducted a human-subject study in which 10 previously untrained operators each completed a set of 3 tasks, 7 times each.  The first task was a plush dice pickup task (30 second timeout), the second was drawer opening (30 second timeout), and last was to place a cup onto a plate (60 second timeout).  The total time for one human subject to learn about the system and complete all tasks took approximately 15 minutes.  Figure~\ref{fig:plots} reports the completion times of each operator on each of the three tasks.

Although the underlying technology is complex, the user interface was easy to understand and use for all operators. Each operator differed in their style of motion, stances, and appearances, but there were no noticeable discrepancies in the behavior of the system or the distribution of results.

We found that subjects often struggled during the first few trials.  However, all subjects found it easy to adjust and learn how to use the system very quickly. Our system was often complimented on its responsiveness and fluidity: subjects did not notice with any lag or jitter in the robot's imitation. Subjects enjoyed participating in the study, and some said that teleoperation of the robot was similar to a video-game.  Additionally, subjects noted they felt safe and comfortable during teleoperation.

The largest frustrations with the system was in periodic errors in the retargeting of the human fingers to the Allegro robot hand.  Many subjects noted instances when they were attempting complicated hand poses, but our system failed to accurately imitate them.  In particular, we noticed systematic errors of our system in handling the flexion of the thumb.  The shape and joint axes of the Allegro hand thumb are particularly different from that of the human thumb, and we suspect that our energy function does not place enough weight on accurate thumb retargeting.  Some subjects observed that the system was worse at tracking their hand when it was all the way open with their palm parallel to the camera, this is a particular issue that we cannot get around with a single camera setup.

\section{Analysis}
\label{sec:analysis}

\subsection{Accuracy of retargeter network}
We compare the accuracy of \texttt{\baseline}'s \textit{online optimization} with our neural network retargeter that relies on \textit{offline optimization} during training.  We gather a test set of 500 sequences from the DexYCB video dataset \cite{chao:cvpr2021}, which contains videos with annotated ground-truth human hand poses.  For each video, at each timestep, the poses are fed to both our neural network and \texttt{\baseline} with a (generous) time budget of 40ms to solve.  We emphasize that both retargeters optimize the same energy function, but in different ways.

We do not, however, have access to ``ground-truth" Allegro joint angles against which to compare the output of the two retargeters. To circumvent this, we design a version of \texttt{\baseline} that is allowed an \textit{infinite time budget} to run until convergence.  We call this the pseudo-ground-truth oracle,  and our assumption is that its final output is as close to optimal as possible.

We compare the root mean squared error (RMSE) between the oracle's outputs and the outputs of each of our two retargeters on the dataset.  Our neural network retargeter outperforms \texttt{\baseline} in matching the oracle. The neural network retargeter achieves an RMSE of \textbf{0.17} radians (about 10 degrees) while \texttt{\baseline} achieves an RMSE of \textbf{0.25} radians per joint (about 14 degrees). 

\begin{figure}[t]
 \centering
 \includegraphics[width=\linewidth]{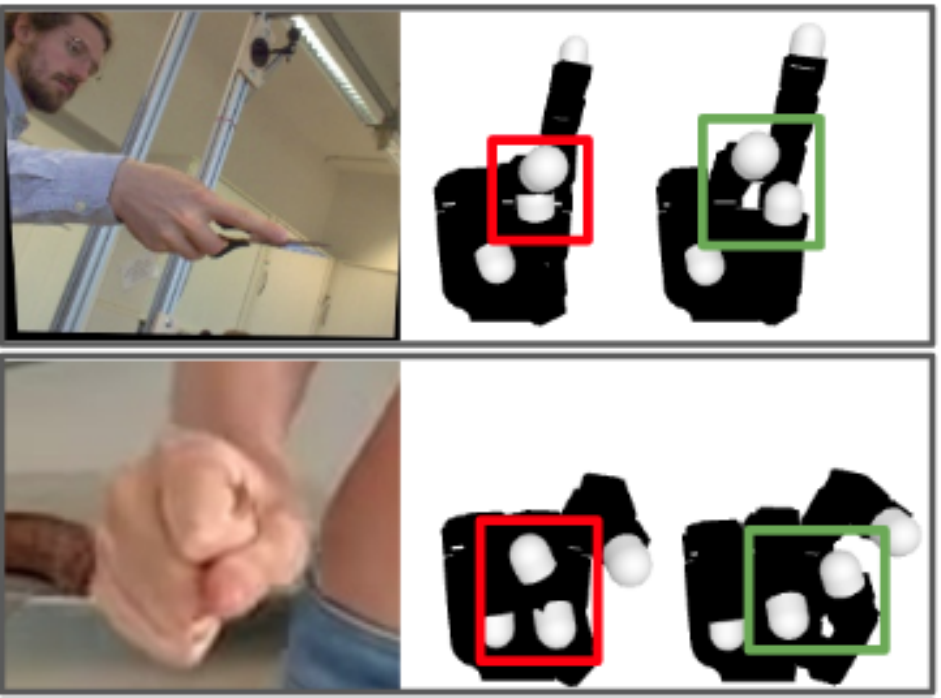}
 \vspace{-0.15in}
 \caption{\small \textbf{The contribution of an adversarial self-collision loss}.  The red boxes highlight instances where the vanilla retargeting network outputs Allegro hand poses that result in self-collision.  The green boxes depict the predictions of the network trained with self-collision loss.  These robot hand poses maintain functional similarity to the human's hand pose, but avoid self-collision.}
 \vspace{-0.05in}
 \label{fig:self_collision}
\end{figure}

\subsection{Self-Collision Avoidance}
We perform an ablation on the weight of the self-collision classifier (Section~\ref{sec:collision}) in the energy function, to see how it affects the behavior of the hand retargeter.  We use a test set of 3000 held-out hand poses from the FreiHand dataset \cite{Freihand2019} and consider 6 different hand retargeter networks, trained with collision-loss weights of 0, 0.2, 0.4, 0.6, 0.8 and 1.  (A weight of 0.8, for example, means that the self-collision loss is weighed 0.8 as heavily as the sum of all the other key-vector matching loss terms in the energy function.)  Each network makes predictions on the data and we compute (1) the fraction of resulting Allegro joint angle vectors that result in self-collision, and (2) the average value of the key-vector energy terms over the dataset.

\begin{figure}[t]
\vspace{-0.15in}
\centering
\includegraphics[width=\linewidth]{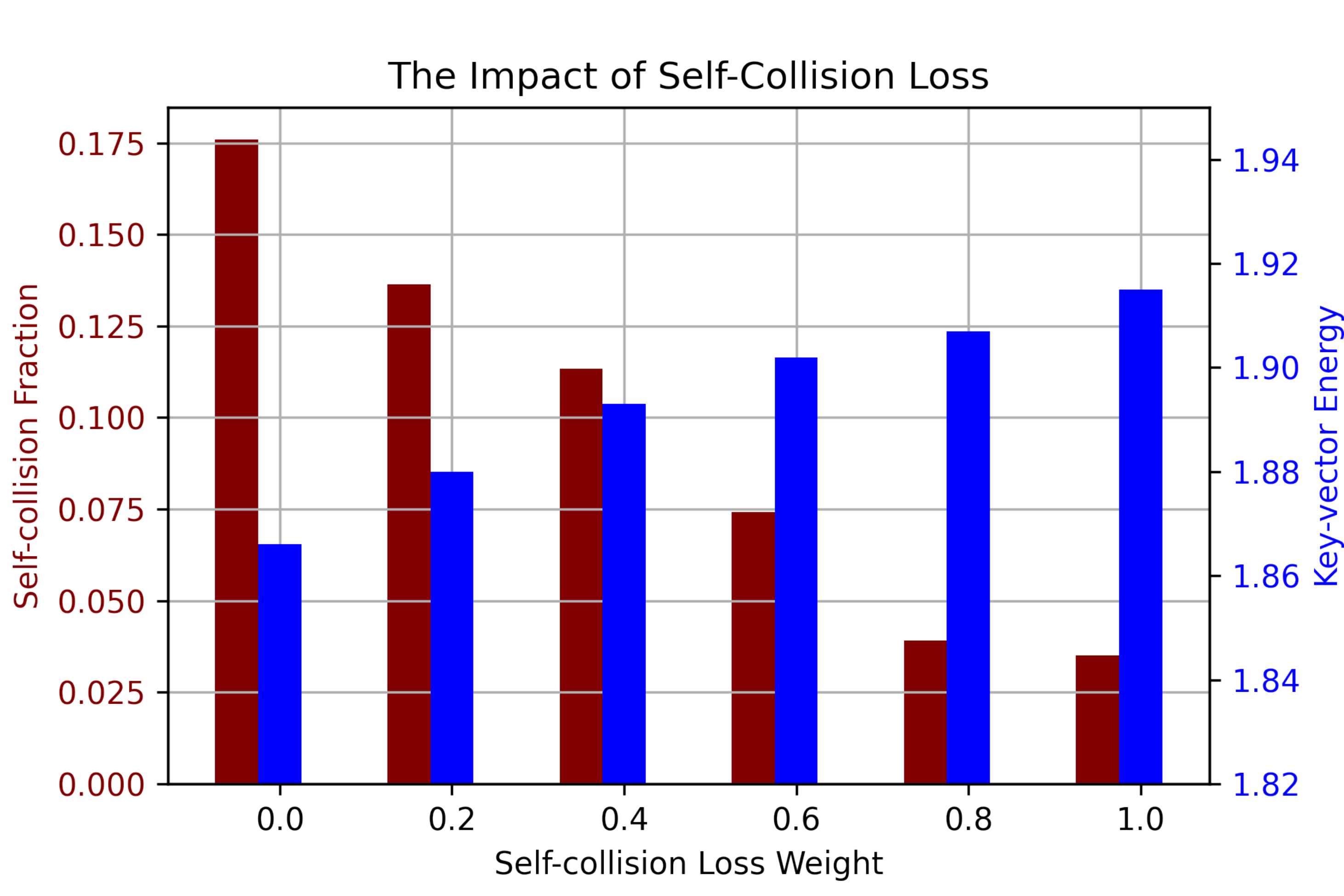}
\vspace{-0.2in}
\caption{\small As the weight of the adversarial self-collision loss is increased, the hand retargeter network produces fewer self-colliding joint configurations (maroon), but incurs a higher energy with less similar poses (blue).  A higher energy means that the predicted robot hand pose is dissimilar to the operator's hand pose.}
\label{fig:coll_vs_energy}
\vspace{-0.11in}
\end{figure}

We summarize the results in Figure~\ref{fig:coll_vs_energy}.  The plot shows there is a trade off between minimizing self-collisions, and minimizing key-vector dissimilarity. As we increase the weighting term of the self-collision avoidance loss term in the energy function, we produce fewer offending joint configurations but minimization performance degrades for the other terms in the energy function.  We depict this trade off visually in Figure~\ref{fig:self_collision}.  It is difficult to confidently assert that one is more valuable than the other, and in practice, we find that a middle ground works very effectively for the user.  

\section{Conclusion}
\label{sec:conclusion}
We present \textit{Robotic Telekinesis}, a system for in-the-wild, real-time, remote visual teleoperation of a dexterous robotic hand and arm, in which a human operator demonstrates tasks to the robot just by moving their own hands.  We leverage the latest advancements in 3D human pose estimation and thousands of hours of raw day-to-day human footage on the internet to train a system that can understand human motion, and retarget it to corresponding robot actions.  Our method requires only a single color camera, and can be used out-of-the box by any operator on any task, without any actively collected robot training data.  We show that our system enables experts and novices alike to successfully perform a number of different dexterous manipulation tasks.  We hope that our system is used as a starting point for future research, rather than as an end product.  We believe a powerful use of visual teleoperation is to bootstrap autonomous robot learning, and by building an intuitive and low-cost platform for humans to provide task demonstrations, we hope to contribute to the democratization of robot learning.

\section*{Acknowledgments}
We are grateful to Ankur Handa for initial feedback and to Shikhar Bahl, Murtaza Dalal and Russell Mendonca for feedback on the paper.  We would also like to thank Murtaza Dalal, Alex Li, Kathryn Chen, Ankit Ramchandani, Ananye Agarwal, Jianren Wang, Zipeng Fu, Aditya Kannan, and Sam Triest for helping with experiments.  KS is supported by NSF Graduate Research Fellowship under Grant No. DGE2140739. The work was supported in part by NSF IIS-2024594, ONR N00014-22-1-2096 and GoodAI Research Award.

\bibliographystyle{plainnat}
\bibliography{main}

\clearpage
%

\begin{appendices}

\section{Hand/Body Bounding Box Detection}

The first step in our retargeting pipeline is to detect two bounding boxes in an image of the human operator; one for the body and one for the right hand.  These bounding boxes needn't be perfectly tight bounding boxes; it's more important that they contain the entire body/hand without truncations.  We use an \href{https://github.com/facebookresearch/frankmocap/blob/main/bodymocap/body_bbox_detector.py}{implementation} of OpenPose \cite{cao2019openpose} from the authors of FrankMocap \cite{FrankMocap_2021_ICCV}.  First, a 2D body skeleton detector is run over the entire image, and outputs the predicted pixel locations for each of the 18 keypoints on the skeleton.  The tight bounding rectangle around the points is then computed, and a fixed padding is applied on all sides to allow a margin of error.

The right hand bounding box is heuristically extracted based on the 2D body skeleton estimate. The bounding box is centered at the pixel corresponding to the right hand wrist, and the side length of the bounding box is conservatively chosen to ensure that the bounding box contains the entire hand.  For an image of size 480x640, we use a side length of 150 pixels.

\section{Hand Pose Estimation}

The next step is to estimate the pose of the operator's right hand from a crop of the hand.  The crop of the right hand is computed as described in the previous section, and is resized to a shape of 224x224.  The crop is passed to a Convolutional Neural Network (CNN), which outputs a low-dimensional representation of the hand configuration.  We use \href{https://github.com/facebookresearch/frankmocap/blob/main/handmocap/hand_mocap_api.py}{an implementation of the hand pose estimation network} from \cite{FrankMocap_2021_ICCV}.  This network uses a ResNet50 trunk \cite{resnet}, followed by a Multi-Layer Perceptron (MLP) regression head, which outputs three relevant parameters of the SMPL-X model \cite{SMPL-X:2019}.  (1) $\beta_{h} \in \mathbb{R}^{10}$ describes the \textit{shape} of the hand (the dimensions of each finger and the palm), (2) $\theta_{h} \in \mathbb{R}^{45}$ describes the \textit{pose} of the hand (how the fingers are arranged) and (3) $\phi_{h} \in \mathbb{R}^3$ describes the global orientation of the hand (how the hand's root coordinate frame is rotated in the image coordinate frame).  The SMPL-X model maps the shape and pose parameters ($\beta_{h}$ and $\theta_{h}$) into a full 3D mesh of the hand, and the global orientation parameter $\phi_{h}$ transforms the coordinate frame of the mesh so the axes align with the axes of the image coordinate frame.

\section{Human-to-Robot Hand Retargeting}

Here, we describe two implementations of the human-to-robot hand retargeting module, one which uses online optimization (via inference-time gradient descent), and one which uses offline optimization (via a neural network).  Both implementations take a human hand pose as input, and outputs joint angles for each of the 16 Allegro hand joints.  The human hand pose is parameterized by the tuple $(\beta_{h}, \theta_{h})$ as described in the previous section.  The global orientation $\phi_{h}$ is not used in hand retargeting, because the Allegro hand has no wrist or palm joints, and therefore, matching the global orientation of the human hand is accomplished by the robot arm and not the robot hand.  We use $q_a$ to denote the vector of the 16 Allegro hand joint angles.

\subsection{Human-to-Robot Hand Energy Function.} Both implementations of the hand retargeting module minimize the same \textit{energy function}, so we describe this first.  Inspired by \cite{dex}, the energy function aims to capture the functional similarity between a human hand pose and a robot hand pose.  Five \textit{keypoints} are defined on each hand: the index fingertip, the middle fingertip, the ring fingertip, the thumb fingertip, and the palm center.  Each of these keypoints is associated with a coordinate frame, and the keypoint is the origin of the coordinate frame.  Enumerating all pairs of keypoints yields ten \textit{keyvectors}.  Four of them are finger-to-palm keyvectors (index-to-palm, middle-to-palm, ring-to-palm and thumb-to-palm), three are inter-finger keyvectors (index-to-middle, index-to-ring and middle-to-ring), and three are finger-to-thumb keyvectors (index-to-thumb, middle-to-thumb and ring-to-thumb).  Notably, each keyvector has one endpoint designated as the origin, and the other as the destination.  The keyvectors are expressed in the coordinate basis of the origin keypoint's coordinate frame.  We refer the reader to Figure 8 of \cite{dex}, which elegantly depicts the keypoint coordinate frames and the keyvectors on both the human and Allegro hand.

The energy function between a human hand pose (parameterized by the tuple $(\beta_{h}, \theta_{h})$) and an Allegro hand pose (parameterized by the joint angles $q_{a}$) is computed as follows.  First, for each $i \in \{1, \dots, 10\}$, the $i$-th keyvector is computed on the human hand (call it $\mathbf{v_i^h}$) and the Allegro hand (call it $\mathbf{v_i^a}$).  Then, each Allegro hand keyvector $\mathbf{v_i^{a}}$ is scaled by a constant $c_i$.  The $i$-th term in the energy function is the Euclidean difference between $\mathbf{v_i^h}$ and $\mathbf{v_i^a}$:
\begin{equation}
    E( \ (\beta_{h}, \theta_{h}), \  q_{a} \ ) = \sum_{i=1}^{10} || \mathbf{v_i^h }- (c_i \cdot \mathbf{v_i^a}) ||_2 ^2
\end{equation}
These scaling constants $\{c_i\}$ are hyperparameters that require some tuning.  If the goal is to produce aesthetically appealing retargeted Allegro hand poses on generic hand gestures, one should set each $c_i$ to around $0.625$, in order to account for the ratio in sizes between the average human hand and the Allegro hand.  If the goal is to maximize functional similarity, in theory, one should set each $c_i$ to 1, to encourage perfect matching of each keyvector.  In practice, we find that setting the constants $c_i$ to a value smaller than 1 is optimal for dexterous manipulation teleoperation.  This is because in order to stably grasp an object, the fingers Allegro hand must exert forces pushing into the object.  This is achieved by commanding the Allegro finger joints to positions that penetrate the object.  These joint angles are never actually reached because the fingers end up colliding with the object, which is precisely the goal.  For our experiments, we use a scaling constant of $0.8$ for each of the finger-to-thumb and finger-to-finger keyvectors, and a scaling constant of $0.5$ for each of the finger-to-palm keyvectors.  This means that in order to ensure a stable grasp, operators must squeeze their fingers closer together than they normally would when grasping, but through our human subject study, we find that novice operators quickly realize this and naturally adjust.

\subsection{Computing the keyvectors on the human hand.}  Having described what the keyvectors are and how they are used to define the energy function, we now describe how to compute the keyvectors on the human hand, given the SMPL-X model parameters $(\beta_{h}, \theta_{h})$.  The first step is to use the SMPL-X model to generate a full posed 3D mesh of the human hand.  Given $\beta_{h}$ and $\theta_{h}$ (and a template hand mesh), the SMPL-X model generates a 3D mesh that correctly captures the shape and pose of the human hand.  The next step is to transform these vertices into a canonical coordinate frame, centered at the palm center, with the positive $x$ axis pointing out of the hand, the positive $y$ axis pointing toward the thumb, and the positive $z$ axis pointing toward the middle fingertip.  This is done by applying a hand-coded transformation between the SMPL-X coordinate frame and the canonical coordinate frame.  The next step is to compute the transformation between each of the keypoint coordinate frames and the canonical coordinate frame.  This is done using the Kabsch-Umeyama Algorithm \cite{umeyama1991least} for estimating the transformation that best aligns corresponding pairs of points.  Concretely, for each keypoint, we manually determine four vertices on the template hand mesh: (1) the keypoint itself, (2) a vertex located along $0.05$m along the positive $x$ axis from the keypoint, (3) a vertex located $0.05$m along the positive $y$ axis from the keypoint, and (4) a vertex located $0.05$m along the positive $z$ axis from the keypoint.  This is pre-computed once, up front.  At runtime, for a given posed human hand mesh, we gather the 3D coordinates for each of these three points in the canonical coordinate frame.  We define a corresponding set of four points: $\{[0, 0, 0], [0.05, 0, 0], [0, 0.5, 0], [0, 0, 0.5]\}$, which denote the coordinates of these points in the coordinate frame of the keypoint.  Given these four correspondences, the Kabsch-Umeyama computes the transformation between the keypoint coordinate frame and the canonical coordinate frame that best aligns these corresponding point pairs.

\subsection{Computing the keyvectors on the Allegro hand.}  Here, we describe how to compute the keyvectors on the Allegro hand, given a vector of Allegro hand joint angles.  The key idea here is to exploit forward kinematics.  The URDF of the Allegro hand defines the kinematic skeleton of the Allegro hand.  The forward kinematics map takes as input a joint angle vector and outputs the transformation between each link's coordinate frame and the root coordinate frame.  Each of our keypoints conveniently corresponds to a particular link on the Allegro hand, so the keypoint coordinate frames can simply be read off from the forward kinematics result.

\subsection{The energy function is differentiable.}  One critical point to note is that the forward kinematics map is a fully differentiable function of the Allegro hand joint angles.  This is because forward kinematics is essentially a chain of sines, cosines and matrix multiplications.  This is important because it means that the energy function is a fully differentiable function of the Allegro joint angles (by the Chain Rule).  This is important because it allows us to compute the gradient of the energy function with respect to the Allegro joint angles, and use gradient descent to find the Allegro joint angles that minimize the energy, with respect to a given human hand pose.  We now describe two ways to exploit this differentiability: via online gradient descent and via offline gradient descent.

\subsection{Retargeting via Online Gradient Descent.}  We implement the online optimization retargeter by using Stochastic Gradient Descent (SGD) to minimize the energy function.  At each iteration, we seed the Allegro joint angles to an initial value.  At the first iteration, that initial value is the all-zero vector (which corresponds to an open palm with outstretched fingers).  At all subsequent iterations, the seed vector is the result from the previous iteration.  We then run a fixed number of SGD steps with a learning rate of $0.05$.  The number of gradient steps is a hyperparameter, and presents a tradeoff between accuracy and speed.  Running more steps results in better convergence, but takes more time.  We find 100 steps to be a good point on this spectrum.

\subsection{Retargeting via Neural Networks.}  We implement the offline optimization retargeter with a neural network that takes as input a human hand pose parameterization $\mathrm{concat}(\beta_{h}, \theta_{h}) \in \mathbb{R}^{55}$ and outputs a vector of Allegro joint angles $q_a \in \mathbb{R}^{16}$.   We use a Multi Layer Perceptron (MLP) with three hidden layers of sizes 256, 256, 128, and intermediate tanh activations.  We apply a tanh to the final output to squeeze the values from $[-\infty, \infty]^{16}$ to $[-1, 1]^{16}$, and then scale the squeezed values to the appropriate values within each of the joint's ranges.  (For example, joint 1 on the Allegro hand has a range of $[-0.196, 1.61]$ (in radians).  If the raw output of the network for this joint were $1.23$, the tanh operation would squeeze it to $0.84$ and the rescale operation would rescale it to $1.47$ radians, which is $0.84$ of the way between $-0.196$ and $1.61$.)

We train this network using a mixture of human hand poses from the Freihand Dataset \cite{Freihand2019} and ``in-the-wild" human hand poses from the 100 Days of Hands dataset \cite{100doh}.  The Freihand dataset contains ground-truth SMPL-X shape and pose parameters for over 30,000 hand configurations, which we use as inputs to the network.  The 100 Days of Hands dataset is simply a list of links to YouTube videos that depict humans using their hands for everyday tasks.  In total, these span hundreds of millions of frames.  We generate human hand poses by running our Hand Pose Estimation module (built on the CNN from \cite{FrankMocap_2021_ICCV}).  Our final dataset consists of 30,000 samples from the FreiHand dataset and 30,000 randomly chosen samples from the 100 Days of Hands dataset.

\section{Body Pose Estimation}

The body pose estimation pipeline consists of two steps.  The first is to estimate a rough body pose from a crop of the operator's body.  The second is to refine the right-hand portion of the rough pose estimate by fusing in the more accurate hand pose estimate from the Hand Pose Estimation module.

\subsection{Rough Body Pose Estimation via a CNN} This step takes as input a crop of the operator's body, resized to a shape of 224 x 224.  The crop is passed to a CNN, which outputs a low-dimensional representation of the body configuration.  We use \href{https://github.com/facebookresearch/frankmocap/blob/main/bodymocap/body_mocap_api.py}{an implementation of the body pose estimation network} from \cite{FrankMocap_2021_ICCV}.  This network uses a ResNet50 trunk \cite{resnet}, followed by a Multi-Layer Perceptron (MLP) regression head, which outputs three relevant parameters of the SMPL-X model.  (1) $\beta_{b} \in \mathbb{R}^{10}$ describes the body \textit{shape} (the dimensions of the various body parts), (2) $\theta_{b} \in \mathbb{R}^{45}$ describes the \textit{pose} of the body (how the limbs are arranged) and (3) $\phi_{b} \in \mathbb{R}^3$ describes the global orientation of the body (how the body's root coordinate frame is rotated in the image coordinate frame).  The pose parameter $\theta_{b}$ is of shape $(24, 3, 3)$, and each of these 24 matrices denotes the 3 x 3 rotation matrix of a particular joint in the human body skeleton.  The SMPL-X model maps the shape and pose parameters ($\beta_{b}$ and $\theta_{b}$) into a full 3D mesh of the body, and the global orientation parameter $\phi_{b}$ transforms the coordinate frame of the mesh so the axes align with the axes of the image coordinate frame.

\subsection{Body-Pose Refinement via Hand Pose Integration.}  The body pose estimate obtained via the CNN can fail to capture the finer details of the hand pose, and crucially, can produce incorrect estimates for the rotation of the right-hand wrist relative to its parent in the human body kinematic chain.  The Hand Pose Estimation module, however, operates on a zoomed-in crop of the operator's hand, and often produces much better estimates of the hand's global orientation.  To exploit this fact, we use the ``Copy-and-Paste" Body-Hand Integration module from \cite{FrankMocap_2021_ICCV}, which refines the local orientation of the wrist based on $\phi_{h}$, the global orientation of the hand estimated by the Hand Pose Estimation module.

\section{Human-to-Robot Body Retargeting}  We now describe how to map from a body pose estimate $(\beta_{b}, \theta_{b})$ to a target pose for the xArm6's end-effector link, relative to its base link.  The first step is to define two pairs of corresponding coordinate frames between the human and robot ``bodies".  The first pair is between the human torso and the robot ``torso".  We define the robot torso to be 25cm above the robot base frame.  Both torso frames are oriented such that the positive $x$ axis points out of the front of the torso, the positive $y$ axis points towards the left side of the body, and the positive $z$ axis points upwards toward the head.  The second pair of coordinate frames is between the human's right hand wrist and the robot's wrist (i.e. end-effector).  The wrist coordinate frames are centered at the wrist center, with the positive $x$ axis oriented parallel to the vector originating at the palm center and pointing out of the front of the palm, the positive $y$ axis pointing toward the thumb, and the positive $z$ axis pointing toward the middle fingertip.

The problem of determining the relative transformation between the robot's end-effector and its base coordinate frame reduces to the problem of determining the relative transformation between the end-effector and the torso (because the torso coordinate frame is fixed relative to the robot's base frame).  And this problem reduces to determining the relative transformation between the human's right hand wrist coordinate frame and the human's torso coordinate frame.  In order to do this, we start at the torso joint, and traverse the human body kinematic chain (defined by the SMPL-X model) from the torso to the wrist, chaining rotations along the path.

\section{xArm6 Inverse Kinematics Controller}
The human-to-robot body retargeter module outputs target poses for the xArm6's end-effector, relative to it's base coordinate frame.  The final step is to build a model that uses this target pose to send a steady and smooth stream of joint angle commands to the xArm6's default controller.  In practice, we found that this module is crucial for performance and must be carefully implemented with attention to details; if the robot arm does not move smoothly, dexterous manipulation tasks become impossible.

The first step is to handle outliers caused by erroneous body pose estimates.  This is done by computing the difference between the arm's current end-effector pose and the end-effector pose output by the retargeting module.  If the difference is greater than a threshold, it is clipped.  The next step is to combine the (possibly clipped) end-effector pose target with a running Exponentially Moving Average (EMA) of end-effector poses.  This helps ensure smooth motion in the presence of noise in the pose estimation and retargeting modules.  The following update rule is used to update running average $P_{EMA}$ to incorporate the new target pose $P_{new}$:
\begin{equation}
    P_{\mathrm{EMA}} = \alpha \cdot P_{\mathrm{new}} + (1 - \alpha) \cdot P_{\mathrm{EMA}}
\end{equation}
We find $\alpha = 0.25$ to work well.  We note that a lower value of $\alpha$ can introduce lag, but we find that because our system runs at such a high frequency, this is not an issue in practice.

The next step is to compute the difference between the robot's current end-effector pose, and the (newly updated) pose target, and to apply linear interpolation to divide that difference into equally spaced waypoints.  Each waypoint end-effector pose is then passed to a Selectively Damped Least Squares (SDLS) Inverse Kinematics (IK) solver \cite{SDLS}, implemented in PyBullet \cite{coumans2020}, which returns a vector of joint angles for the six joints in the xArm6.  These joint angle commands are sent to the xArm6 servo controller. 

\section{Software Architecture}

We now describe how we put together all of the aforementioned modules into a single system that efficiently retargets human motion to robot trajectories.  We found it natural to design our system as a dataflow graph, with computation being done at the nodes, and inputs/outputs travelling along the edges.  We first summarize the computation nodes we use, and then discuss how we optimized runtime performance by using parallel computation within a publisher-subscriber architecture.

\subsection{The nodes in the dataflow graph.}  Each node corresponds roughly to one of the modules described in previous sections:
\begin{itemize}
    \item \textbf{CameraNode}: captures RGB images of the operator at 30Hz.
    \item \textbf{HandBoundingBoxDetectorNode}: receives an operator image, and computes a bounding box of the right hand.
    \item \textbf{BodyBoundingBoxDetectorNode}: receives an operator image, and computes a bounding box of the body.
    \item \textbf{HandPoseEstimationNode}: receives a crop of the operator's right hand, and estimates the SMPL-X model parameters $(\beta_{h}, \theta_{h}, \phi_{h})$ that parameterize the hand's shape, pose and global orientation.
    \item \textbf{BodyPoseEstimationNode}: receives a crop of the operator's body, and estimates the SMPL-X model parameters $(\beta_{b}, \theta_{b}, \phi_{b})$ that parameterize the body's shape, pose and global orientation.
    \item \textbf{BodyHandIntegrationNode}: receives a hand pose estimate $(\beta_{h}, \theta_{h}, \phi_{h})$ and a body pose estimate $(\beta_{b}, \theta_{b}, \phi_{b})$, and computes a refined body pose estimate by using the Copy-and-Paste integration method from \cite{FrankMocap_2021_ICCV}.
    \item \textbf{HandRetargetNode}: receives a hand pose estimate $(\beta_{h}, \theta_{h}, \phi_{h})$, and computes the Allegro joint angles $q_a$ that maximizes similarity with the operator's hand.
    \item \textbf{BodyRetargetNode}: receives a (refined) body pose estimate $(\beta_{b}, \theta_{b}, \phi_{b})$, computes the relative transformation between the right hand wrist and the torso, and converts this to a target pose of the xArm6's end-effector link, relative to its base link.
    \item \textbf{AllegroHandControllerNode}: receives a target Allegro hand joint angle vector from the \textbf{HandRetargetNode}, interpolates the difference between robot's current joint angle values and the target into small fixed-size intervals, and sends a stream of interpolated joint angle commands to the robot's controller at a fixed frequency.
    \item \textbf{xArm6ControllerNode}: receives a target end-effector pose from the \textbf{BodyRetargetNode}, and commands a smoothly interpolated stream of xArm6 joint angle configurations to the robot's controller at a fixed frequency.

\end{itemize}

\subsection{Optimizing performance via parallel computation.}
It is crucial that our system run as fast as possible, in order to ensure smooth robot motion and to avoid lagging behind the operator.  Therefore, a key design decision was to opt for a parallel computation paradigm.  The first implementation of our system sequentially chained together the various modules, and achieved a runtime of approximately 3Hz.  In this squential implementation, the retargeting time was the sum of the time taken by each module.  Our optimized implementation instead used the ROS publisher-subscriber architecture, with each node running on a separate process.  Nodes pass inputs and outputs to each other via inter-process messages.  With this approach, the retargeting time was determined only by the slowest node, and this achieved a runtime of approximately 25Hz, which greatly improves usability.

\section{Hardware Architecture}
Our setup consists of a Ufactory xArm6 robot arm mounted to a Vention table, with a Wonik Robotics Allegro Hand mounted as the end-effector. The Allegro Hand was upgraded with four 3D printed fingertips that are skinnier than the default tips. 3M TB641 grip tape is applied to the inner parts of the hand and around the fingertip which allows the Allegro Hand to better grip objects, as 3D printed components and the built in metal/plastic parts are slippery.  One Intel Realsense D415 camera tracks the operator; we use only the RGB stream.  In our experiments, the operator is standing near the robot, but this is not a requirement.  The operator only needs to be able to see the robot, in order for them to adjust their movements to effectively complete tasks.  In the future, we hope to enable this via internet webcams which would allow the operator to be located anywhere in the world.  Running the system is a desktop system with an AMD Ryzen 3960x CPU, 128GB of RAM and two NVIDIA GeForce RTX 3080TI GPU's.

\section{Human Subject Study Details}

The 10 subjects that participated in the study were volunteer colleagues from the author's lab.  A few were familiar with this project, but they were not intimately familiar with the details.  Critically, they had never used the system before.  The human subjects were assured that the data collected was anonymous, the robot never interacted with them in any way, and if they ever felt uncomfortable with the task for any reason they could terminate the experiment early.  The act of collecting the data would fall under a Benign Behavioral Intervention:  verbal, written responses, (including data entry or audiovisual recording) from adult subjects who prospectively agrees and the following is met: Recorded information cannot readily identify the subject (directly or indirectly/linked).   This therefore gives an exemption for IRB approval.  Example of this category are solving puzzles under various noise conditions, playing an economic game, being exposed to stimuli such as color, light or sound (at safe levels), performing cognitive tasks.

One author was the conductor of the study.  The conductor briefed each subject on how to operate the system.  The subjects were asked to stand and stay in frame of the camera during the duration of the experiments.  They were asked to not move around too quickly as that would trigger safety limits of the control system, but this was never an issue.  No other significant instructions were given.  The conductor of the study also kept an emergency stop switch next to them for the safety of the robot system, but it was never used.  

For the first few trials, many subjects were confused by the system but quickly adapted to it.  The conductor instructed them to continually adapt to the system and try to complete the tasks without giving them additional information.  The conductor took down notes on the compliments and complaints of the system while the system was being used.

The conductor only verbally told each subject the goal of the task but did not explain the best way to complete them.  The tasks were very simple and intuitive so the subjects were not confused by them.  For each task, a failure was recorded when either the time expired, the task became impossible to complete from the object state on the table, or the subject asked for a reset.  Between each trial within each task, the subjects were asked to move the robot arm up away from the table to allow the conductor to reset the object.  Between each task, the Telekinesis system was paused and subjects were allowed to rest their arm for a few minutes.  The dice pickup task and drawer task was 30 seconds each for 7 trials.  The last cup in plate task was 60 seconds long for each of the 7 trials.  The total time to complete all three tasks was about 15 minutes.

\end{appendices}

\end{document}